\documentclass[10pt, a4paper]{article}
\usepackage{lrec2022} % this is the new LREC2022 Style
\usepackage{multibib}
\newcites{languageresource}{Language Resources}
\usepackage{graphicx}
\usepackage{tabularx}
\usepackage{soul}
\usepackage{titlesec}
\titleformat{\section}{\normalfont\large\bfseries\center}{\thesection.}{1em}{}
\titleformat{\subsection}{\normalfont\SmallTitleFont\bfseries\raggedright}{\thesubsection.}{1em}{}
\titleformat{\subsubsection}{\normalfont\normalsize\bfseries\raggedright}{\thesubsubsection.}{1em}{}
\renewcommand\thesection{\arabic{section}}
\renewcommand\thesubsection{\thesection.\arabic{subsection}}
\renewcommand\thesubsubsection{\thesubsection.\arabic{subsubsection}}
\usepackage{epstopdf}
\usepackage[utf8]{inputenc}

\usepackage{hyperref}
\usepackage{xstring}

\usepackage{color}

\title{Towards Understanding Gender-Seniority Compound Bias\\in Natural Language Generation}

\name{%
\begin{tabular}{@{}c@{}}
Samhita Honnavalli$\ast^1$\thanks{$\ast$ Equal contribution}, 
Aesha Parekh$\ast^1$,  
Lily Ou$\ast^1$,  
Sophie Groenwold$\ast^1$,\\ 
Sharon Levy$^1$,  
Vicente Ordonez$^2$,  
William Yang Wang$^1$
\end{tabular}}
\address{
\textsuperscript{1}Department of Computer Science, University of California Santa Barbara \\
\textsuperscript{2}Department of Computer Science, Rice University \\
\{shonnavalli, aeshaparekh, lilyou, sophiegroenwold\}@ucsb.edu \\ 
\{sharonlevy,william\}@cs.ucsb.edu \\ 
vicenteor@rice.edu}

\abstract{
Women are often perceived as junior to their male counterparts, even within the same job titles. While there has been significant progress in the evaluation of gender bias in natural language processing (NLP), existing studies seldom investigate how biases toward gender groups change when compounded with other societal biases. In this work, we investigate how seniority impacts the degree of gender bias exhibited in pretrained neural generation models by introducing a novel framework for probing compound bias. We contribute a benchmark robustness-testing dataset spanning two domains, U.S. senatorship and professorship, created using a distant-supervision method. Our dataset includes human-written text with underlying ground truth and paired counterfactuals. We then examine GPT-2 perplexity and the frequency of gendered language in generated text. Our results show that GPT-2 amplifies bias by considering women as junior and men as senior more often than the ground truth in both domains. These results suggest that NLP applications built using GPT-2 may harm women in professional capacities.
 \\ \newline  }

\begin{document}

\maketitleabstract

\section{Introduction}
Propagation of societal biases is a growing issue in mainstream natural language generation (NLG) models. Downstream applications of these models, such as machine translation \cite{statistical-machine-translation}, dialogue generation \cite{dialogue-gen}, and story generation \cite{yao2019plan} risk reinforcing societal stereotypes.

One of the most well-known types of societal bias in natural language processing (NLP) is gender bias \cite{sun-etal-2019-mitigating,zhao-etal-2019-gender,bolukbasi2016man,rudinger-etal-2018-gender}. Previous work has revealed gender bias in coreference systems using an evaluation corpus that links gendered entities to various occupations \cite{zhao-etal-2018-gender}. Similarly, \newcite{kurita-etal-2019-measuring} quantifies gender bias using probabilities that BERT \cite{devlin-etal-2019-bert} assigns to sentences that associate gendered words with career-related words. Although the impact of gender bias on NLP tasks has been consistently identified and measured \cite{zhao-etal-2017-men,Bordia_2019}, we hypothesize that it does not occur in isolation. In this paper, we view bias through a multidimensional lens by studying compound gender-seniority bias.

Due to gender stereotypes, traits typically associated with high-seniority positions, such as leaders in a given field, are more often attributed to men than to women \cite{eagly-karau-role-congruity,heilman-gender-stereotypes}. Consequently, natural language generation (NLG) models may be perpetuating biased information about gendered entities with respect to their perceived seniority level. We have seen how bias in NLP has disproportionately harmed already-marginalized communities through the use of downstream applications before -- for example, when companies and universities have sought to apply or actively used NLP for applicant-filtering systems. These use cases in particular can prevent qualified women from having the same professional opportunities as men. Seniority has the potential to influence and exacerbate gender bias in real-world systems that utilize NLP: human resources chatbots and resume scanning systems deal with both seniority and gender. Using gender- or seniority-biased models in sensitive applications of NLP can potentially worsen the existing representation gap, so as a first step it is important to identify where these biases occur.

\begin{table}[t!]
\begin{minipage}{0.47\textwidth}
\small
\centering
\setlength{\tabcolsep}{5.5pt}
\newcommand{\centered}[1]{\begin{tabular}{l} #1 \end{tabular}}
\begin{tabular}{|@{}l@{}|@{}l@{}|}
% \begin{tabular}{|c|l|}
    \hline
    \centered{\\ \textsc{Original} \\\\} & 
    \centered{Our \textbf{junior} Senator \textbf{Shelley Moore} \\
     \textbf{Capito} sits on this important \\
     committee... \\}
    \\ \hline
    \centered{\textsc{flip by } \\ \textsc{seniority}} &  
    \centered {Our \textbf{senior} Senator Shelley Moore \\
    Capito sits on this important \\ committee...} 
    \\ \hline
    \centered{\textsc{flip by} \\ \textsc{gender}} &
    \centered {Our junior Senator \textbf{Tom Cotton} sits \\ on
      this important committee...}
    \\ \hline
\end{tabular}
\caption{An example of an original human-written sample and its counterfactuals from the U.S. Senate domain in our corpus. The phrase acts as a prompt for the perplexity experiment. Flipped entities are in bold.}
\label{fig:true-false-example}
\end{minipage}    
\vspace{-2ex}
\end{table}

To determine the extent to which seniority affects the bias in current NLG systems, we perform a systematic study of gender and seniority bias in GPT-2 \cite{radford2019language}, a Transformer-based language model, across two domains: the U.S. Senate and U.S. university professors. To examine the bias resulting from the compound of gender and seniority, we create a distantly-supervised dataset of human-written samples from Google search results. We adopt a distant supervision method for high-precision sample collection, an example of which can be seen in Table \ref{fig:true-false-example}.

We conduct two experiments: one to observe the gender-seniority compound bias, and another to demonstrate the impact of seniority on gender bias. These experiments indicate that seniority significantly influences gender bias in GPT-2, demonstrating that women have a higher association with junior rankings and men have higher association with senior rankings in both domains we study. This in turn amplifies both representation and promotion bias for women in professional spheres. Our contributions include:

\begin{itemize}
     \item A novel, multi-factor framework for investigating gender and seniority bias in pretrained generative models.
     \item A high-precision dataset spanning two domains, collected by distant-supervision methods, which can be used to build robust NLG models in future work.\footnotemark[1]
     \item An identification and analysis of GPT-2's association of women with junior positions and men with senior positions using our dataset, demonstrating amplified bias.  
\end{itemize}  

\footnotetext[1]{https://github.com/aeshapar/gender-seniority-compound-bias-dataset}

\section{Domains}

To investigate the gender-seniority bias, we look to two domains with well-defined notions of seniority: the U.S. Senate and U.S. professors. For each domain, we gather the names of those with available gender and seniority labels: the 2020 U.S. Senate ($n = 100$) and a set of professors from the the 2014 U.S. News top 50 U.S. Computer Science graduate programs ($n = 2220$) \cite{Papoutsaki2015CrowdsourcingFS}. 

\begin{table}[!t]
\begin{minipage}{0.47\textwidth}
\small
\centering
\setlength{\tabcolsep}{5.5pt}
\newcommand{\centered}[1]{\begin{tabular}{l} #1 \end{tabular}}
\begin{tabular}{|r|cc|cc|}
\hline
    & \multicolumn{2}{c|}{\textbf{Senators}} & \multicolumn{2}{c|}{\textbf{Professors}} 
    \\
    \cline{2-5}
    & \textit{Female} & \textit{Male} 
    & \textit{Female} & \textit{Male}
    \\ \hline
    \textit{Junior/Assistant} & 225 & 562 & 1064 & 1018 \\
    \textit{Senior/Associate} & 179 & 598 & 1064 & 1033 \\ 
\hline
\end{tabular}
\caption{Original, validated sample counts for Senators and professors, by seniority and gender classes.}
\label{fig:senator-prof-dataset}
\end{minipage}    
\vspace{-2ex}
\end{table}

Seniority in these domains is defined as follows. Each U.S. state has two senators, where the senator with the longer incumbency is the senior senator for that state and the other is the junior senator. Most professors in U.S. universities fall into one of three seniority categories: (from least senior to most) assistant, associate, and full professors.

\renewcommand{\arraystretch}{1.15} % number is multiplier of default line height
\begin{table*}[t]
\small
\centering
\setlength{\tabcolsep}{5.5pt}
\begin{tabular}{|cr|cccc|cccc|}
    \hline 
    & &  \multicolumn{4}{c|}{\textbf{Senators}} & \multicolumn{4}{c|}{\textbf{Professors}}
    \\ \cline{3-10}
    & & \textit{Jr. Female} & \textit{Jr. Male} & \textit{Sr. Female} & \textit{Sr. Male} & \textit{Jr. Female} & \textit{Jr. Male} & \textit{Sr. Female} & \textit{Sr. Male}
    \\ \hline
    \multicolumn{1}{|c|}{} & \textit{Original} & 60.99 & 63.79 & 48.04 & 54.72 & 79.25 & 73.52 & 78.05 & 78.87
    \\
    \multicolumn{1}{|c|}{\textbf{Gender}} & \textit{Flipped} & 71.66 & 72.54 & 62.29 & 62.48 & 79.65 & 80.09 & 79.52 & 85.75
    \\ 
    \multicolumn{1}{|c|}{} & \textit{Delta} & 10.67 & 8.75 & 14.25 & 7.76 & 0.4 & 6.57 & 1.47 & 6.88
    \\
    \multicolumn{1}{|c|}{} & \textit{p-value} & $<$0.01 & $<$0.01 & $<$0.01 & $<$0.01 & 0.236 & $<$0.01 & 0.245 & $<$0.01
    \\ \hline
    \multicolumn{1}{|c|}{} & \textit{Original} & 60.99 & 63.79 & 48.04 & 54.72 & 79.25 & 73.52 & 78.05 & 78.87
    \\
    \multicolumn{1}{|c|}{\textbf{Seniority}} & \textit{Flipped} & 61.38 & 63.09 & 48.79 & 56.41 & 78.08 & 72.76 & 80.03 & 80.48
    \\
    \multicolumn{1}{|c|}{} & \textit{Delta} & 0.39 & -0.7 & 0.75 & 1.69 & -1.17 & -0.76 & 1.98 & 1.61
    \\
    \multicolumn{1}{|c|}{} & \textit{p-value} & 0.153 & 0.034 & $<$0.01 & $<$0.01 & 0.268 & 0.379 & $<$0.01 & 0.003
    \\ \hline
\end{tabular}
\caption{Average perplexity for each gender-seniority class across both U.S. Senator and Professorship domains. Each original-flipped example refers to the original statement and its gender-flipped or seniority-flipped counterfactuals. The Delta denotes the difference in perplexity going from flipped to original. P-values are computed using a Wilcoxon rank-sum significance test.}
\label{fig:perplexity-results}
\vspace{-3ex}
\end{table*}

\section{Distantly-Supervised Dataset Creation}
\label{dataset-creation}

Prior work has utilized distant supervision for relation extraction tasks, where an existing database of relation instances is used to generate large-scale labeled training data \cite{mintz-etal-2009-distant,zeng-etal-2015-distant}. We adopt this method for collecting samples to create datasets for our domains, validate our samples through Amazon Mechanical Turk (AMT), and utilize gender- and seniority-swapping to create paired counterfactuals.

\paragraph{Sample Collection}
To create our dataset, we use high-precision, top-$k$ distantly-supervised Google search results by querying individuals by their full name and seniority standing. For example, senior senator Elizabeth Warren is queried as ``senior senator" ``Elizabeth Warren." Utilizing quotation marks ensures that the name and/or seniority standing appear in the search results. We equate assistant and associate professors to junior and senior ranks, respectively,
because the designation of ``full professor" is often shortened to ``professor," which would be conflated with queries ``assistant professor" and ``associate professor." 
We obtain snippets displayed under each search result: for senators, we use the first two pages of search results, and for professors, just the first (as senators garner a larger number of relevant results). These snippets are then categorized by the individual's gender (which is constrained to binary by our domains) and seniority, giving us four gender-seniority classes: senior/associate female, senior/associate male, junior/assistant female, and junior/assistant male. 

\paragraph{Human Validation}
To ensure the quality of our samples, we employed AMT annotators based in the U.S. with an approval rating of 98\% or above. Annotators were given a query sample and asked to confirm whether it contained the name of the individual queried and their seniority classification. We release a corpus with the validated samples, the statistics for which can be found in Table \ref{fig:senator-prof-dataset}.

\paragraph{Counterfactual Samples}
For each gender-seniority class, we create counterfactual samples to accompany each queried statement using gender swapping procedures \cite{Lu2018GenderBI,kiritchenko-mohammad-2018-examining} as seen in Table \ref{fig:true-false-example}. To seniority swap, we label the queried samples as original statements, then switch the instances of the word ``junior" with ``senior" for senators and ``assistant" with ``associate" for professors, and vice versa in each sample. Likewise, to generate the original-flipped pairs with respect to gender, we utilize the same original statements and swap each instance of male pronouns with female pronouns and a male individual's first and/or last name with a randomly selected first and/or last name from a female individual of the same seniority. The same is done from female to male. 

\section{Ethical Considerations for Dataset Creation}
\paragraph{AMT Compensation}
Regardless of whether the annotated sample was later used in experimentation, all AMT workers were compensated fairly according to U.S. federal minimum wage guidelines: $\$10$ per hour.

\paragraph{Dataset Notes}
We created our dataset taking the top results from Google search, which consist mainly of news articles written in Standard American English. We also used domains specific to the U.S. (American senatorship and professorship), so our samples may reflect societal standards from this country. The decision to study professorship specifically in the field of Computer Science was due to dataset availability.

\paragraph{Intellectual Property}
Google moderates their search results in compliance with the Digital Millennium Copyright Act (DMCA). Thus, any sample collected for our dataset upholds Google's standard for intellectual property rights.\footnotemark[2]
\footnotetext[2]{https://transparencyreport.google.com/copyright/overview}

\paragraph{Gender}
We do not gender individuals ourselves, and instead, the genders associated with the individuals provided by the datasets are used. We acknowledge that unfortunately, we do not study non-binary genders due to the lack of representation in the U.S. Senate and the limited availability of non-binary data for professors. We encourage future work to investigate outside of the gender binary. While we chose senatorship and professorship for their well-defined notions of seniority in this work, future research can be more inclusive in this regard by investigating a domain with higher instances of non-binary individuals.

\section{Quantifying Compound Bias with Perplexity} \label{quant-bias}

To quantify GPT-2's gender-seniority associations, we use GPT-2 Large to compute our dataset's perplexity. The perplexity of a language model is the inverse probability of the test set given the model. Thus, higher perplexity means that GPT-2 finds the sentence less probable and vice versa. We calculate the perplexity of our original-flipped examples across both domains. We downsample each gender-seniority class for balanced classes, yielding $n = 179$ samples for each senator class and $n = 1018$ for each professor class. We include the average perplexity of each class and the results from a Wilcoxon rank-sum significance test in Table \ref{fig:perplexity-results}.

We observe that gender-flipping female to male in the professor domain does not affect the perplexity score, whereas male to female significantly increases its perplexity (see Table \ref{fig:perplexity-results}). This indicates that GPT-2 has a lower propensity to associate female professors with the same rank as male professors, whereas the reverse is not true. Furthermore, the perplexity score increase is slightly larger when going from associate male professor to female than from assistant male to female. This is slightly different with the senator domain because senators are typically prominent figures, belonging to a spectrum within the head distribution, whereas most professors are relatively unknown, and their names are in the long-tail distributions. Gender flipping for professors replaces female names with male names in the same position in the long tail; for senators, results vary by their recognition. Overall these results suggest that there is bias in GPT-2 against female entities and that this bias is greater in association with associate professorships than assistant professorships.

Flipping the seniority in a sentence from assistant to associate decreases its perplexity, whereas flipping from associate to assistant increases it as GPT-2 considers being an associate professor more probable for both male and female individuals.

Additionally, for senator samples,  we notice that the perplexity of female samples increases when we flip from junior to senior, whereas it decreases when we do so for male samples (See Table \ref{fig:perplexity-results}). This reveals that GPT-2 is inclined to consider junior male senators more probable as senior senators, whereas the opposite is true for junior female senators. There is also a greater increase in perplexity when we flip from senior to junior for male samples than for female samples, indicating that GPT-2 is more inclined to associate a junior rank with senior female senators than with senior male senators.
\\
By computing the perplexity of GPT-2 across U.S. professorship and senatorship, we quantify its gender-seniority compound bias and demonstrate a strong association between seniority and gender.

\section{Impact of Seniority on the Frequency of Gendered Language}\label{gendered-bias}
To measure how seniority impacts gender bias in GPT-2, we compare the ground truth distribution of gender to the observed distribution of gendered language in generated text as prompted by phrases where seniority is varied independently. The ground truth ratios for senators correspond to the gender distribution of 2020 U.S. senators, and for professors, they correspond to the data taken from the 2019 Computing Research Association (CRA) Taulbee survey.\footnotemark[3]

We prompt GPT-2 at a temperature of 1, with 3 sets of 10 prompts, for 50 iterations each. Each set contains intent-equivalent gender-neutral prompts, but varied information regarding seniority (See Table \ref{fig:prompt-ex}). Prompts in set 1 do not contain any seniority information, serving as a baseline; set 2 prompts are identical to set 1, except mentions of ``senator" are replaced with ``junior senator"; similarly, for set 3 prompts, mentions of ``senator" are replaced with ``senior senator." We do the same for professors, but with professorship ranks. 
\begin{table}[t]
\begin{minipage}{0.47\textwidth}
\small
\centering
\setlength{\tabcolsep}{5.5pt}
\begin{tabular}{|l|l|}
\hline
\multicolumn{1}{|c|}{Prompt} & \multicolumn{1}{c|}{Generated Text Samples}  \\ \hline
The \textbf{senator} is            & \begin{tabular}[c]{@{}l@{}}expected to announce his\\ known for his progressive views\end{tabular}        \\ \hline
The \textbf{junior senator} is     & \begin{tabular}[c]{@{}l@{}}the first in his family to attend\\ trying to distance himself from\end{tabular}  \\ \hline
The \textbf{senior senator} is     & \begin{tabular}[c]{@{}l@{}}in Washington preparing for her\\ being investigated for his role\end{tabular} \\ \hline
\end{tabular}
\caption{An example of how the seniority for a prompt was varied between the three sets.}
\label{fig:prompt-ex}
\end{minipage}    
\vspace{-2ex}
\end{table}

Through AMT evaluation, we obtain classifications of the gender (with respect to the subject of the sentence) present in the generated texts. The annotators were provided with the generated segments and asked to identify each as containing female-gendered language, male-gendered language, both, or neither. 
%Annotator guidelines are included in Supplementary Materials. 
Results are shown in Table \ref{fig:gendered-results}. 

\footnotetext[3]{https://cra.org/wp-content/uploads/2020/05/2019-Taulbee-Survey.pdf}

For all senator prompts, the percent of male-gendered language in the generated text is greater than the ground truth, whereas the percent of female-gendered language is less than the ground truth. We use a two-sample z-test for each ground truth-observed value pair and find that all pairs are significant with $\alpha = 0.05$ except for male senior senators ($p = 0.06$), male junior senators ($p = 0.14$), female senior senators ($p = 0.06$), and female junior senators ($p = 0.14$). This increased gap between the amount of female and male-gendered language in the generated text indicates an amplification of the representation bias in the U.S. Senate.

If seniority has no influence on gender bias we would expect all the observed junior, senior, and seniority-neutral results to display similar ratios of female to male gendered language. However, the results in Table \ref{fig:gendered-results} reveal that specifying ``junior" causes the model to predict female-gendered text 7\% more often than when seniority is not specified. Prompting GPT-2 with ``senior" causes the model to predict female-gendered text 1.4\% less often and male-gendered text 1.4\% more often than non-specified seniority. This indicates that seniority amplifies the gender bias of GPT-2.

Additionally, for both the assistant and associate professor prompts, we notice that GPT-2 overestimates the proportion of female computer science professors in comparison to the ground truth, which demonstrates an amplification of promotional bias in the field. GPT-2's increased perception of females as assistant professors from ground truth (+18.5\%) is greater than its increased perception of associate professors (+11.5\%). The model also generates 8.3\% more female-gendered language when prompted with ``assistant" than when prompted with ``associate." These results are consistent with the compound bias observed for the senator domain, where females are more often associated with junior positions than senior positions, whereas the opposite is true for males. 

\begin{table}
\small
\begin{center}
\begin{tabular}{ |c|c|c|c|c| }
 \hline
 & \multicolumn{2}{ |c| }{\textit{Male}} & \multicolumn{2}{ |c| }{\textit{Female}} \\
 \cline{2-5}
 & GT & OBS & GT & OBS \\
 \hline
 \textit{Sen.} & 74.0\% & 83.5\% & 26.0\% & 16.5\% \\ 
 \textit{Junior Sen.} & 70.0\% & 76.5\% & 30.0\% & 23.5\% \\ 
 \textit{Senior Sen.} & 78.0\% & 84.9\% & 22.0\% & 15.1\% \\ 
 \hline
 \textit{Prof.} & 77.4\% & 84.2\% & 22.6\% & 15.8\% \\ 
 \textit{Assistant Prof.} & 76.1\% & 57.6\% & 23.9\% & 42.4\% \\ 
 \textit{Associate Prof.} & 77.4\% & 65.9\% & 22.6\% & 34.1\% \\ 
 \hline
\end{tabular}
\caption{Comparison of ground truth (GT) distribution of gender to observed (OBS) distribution of gendered language in GPT-2 generated text for U.S. Senators and U.S. Computer Science Professors.\protect\footnotemark[3]}
\label{fig:gendered-results}
\end{center}
\end{table}

It is difficult to identify the source of bias without access to GPT-2's training data. If the bias is from the data, it could be addressed by also training GPT-2 on a gender- and seniority-flipped dataset. If algorithmic, techniques of algorithm modification, such as \newcite{zhao-etal-2017-men}'s Reducing Bias Amplification conditional model, could be applied.

\section{Conclusion}

By examining perplexity and the frequency of gendered language, we highlight the amplification of gender bias in GPT-2 when compounded with seniority. We create a distantly-supervised dataset across two domains which can be used as a benchmark dataset in future work. We then use the two aforementioned experiments to show that GPT-2 associates senior/associate positions with males and junior/assistant positions with females for both U.S. Senators and professors. Our novel framework can be used for probing other pretrained neural generation models to further investigate compound biases. We hope our findings and methodology can serve as an early intervention to the propagation of these biases, thus decreasing bias-induced harms in downstream applications.

\section{Acknowledgements}
This work was supported by the National Science Foundation award \#2048122 and \#1821415. The views expressed are those of the authors and do not reflect the official policy or position of the U.S. government.

\section{References}
\bibliographystyle{lrec2022-bib}
\bibliography{lrec2022}

\begin{thebibliography}{}

\bibitem[\protect\citename{Bolukbasi \bgroup et al.\egroup
  }2016]{bolukbasi2016man}
Bolukbasi, T., Chang, K.-W., Zou, J., Saligrama, V., and Kalai, A.~T.
\newblock (2016).
\newblock Man is to computer programmer as woman is to homemaker? debiasing
  word embeddings.
\newblock In {\em NIPS}, June.

\bibitem[\protect\citename{Bordia and Bowman}2019]{Bordia_2019}
Bordia, S. and Bowman, S.~R.
\newblock (2019).
\newblock Identifying and reducing gender bias in word-level language models.
\newblock {\em Proceedings of the 2019 Conference of the North}.

\bibitem[\protect\citename{Devlin \bgroup et al.\egroup
  }2019]{devlin-etal-2019-bert}
Devlin, J., Chang, M.-W., Lee, K., and Toutanova, K.
\newblock (2019).
\newblock {BERT}: Pre-training of deep bidirectional transformers for language
  understanding.
\newblock In {\em Proceedings of the 2019 Conference of the North {A}merican
  Chapter of the Association for Computational Linguistics: Human Language
  Technologies, Volume 1 (Long and Short Papers)}, pages 4171--4186,
  Minneapolis, Minnesota, June. Association for Computational Linguistics.

\bibitem[\protect\citename{Eagly and Karau}2002]{eagly-karau-role-congruity}
Eagly, A.~H. and Karau, S.~J.
\newblock (2002).
\newblock Role congruity theory of prejudice toward female leaders.
\newblock page 573–598.

\bibitem[\protect\citename{Heilman}2012]{heilman-gender-stereotypes}
Heilman, M.
\newblock (2012).
\newblock Gender stereotypes and workplace bias.
\newblock {\em Research in Organizational Behavior}, 32:113–135, 12.

\bibitem[\protect\citename{Kiritchenko and
  Mohammad}2018]{kiritchenko-mohammad-2018-examining}
Kiritchenko, S. and Mohammad, S.
\newblock (2018).
\newblock Examining gender and race bias in two hundred sentiment analysis
  systems.
\newblock In {\em Proceedings of the Seventh Joint Conference on Lexical and
  Computational Semantics}, pages 43--53, New Orleans, Louisiana, June.
  Association for Computational Linguistics.

\bibitem[\protect\citename{Koehn}2009]{statistical-machine-translation}
Koehn, P.
\newblock (2009).
\newblock {\em Statistical machine translation}.
\newblock Cambridge University Press.

\bibitem[\protect\citename{Kurita \bgroup et al.\egroup
  }2019]{kurita-etal-2019-measuring}
Kurita, K., Vyas, N., Pareek, A., Black, A.~W., and Tsvetkov, Y.
\newblock (2019).
\newblock Measuring bias in contextualized word representations.
\newblock In {\em Proceedings of the First Workshop on Gender Bias in Natural
  Language Processing}, pages 166--172, Florence, Italy, August. Association
  for Computational Linguistics.

\bibitem[\protect\citename{Lu \bgroup et al.\egroup }2018]{Lu2018GenderBI}
Lu, K., Mardziel, P., Wu, F., Amancharla, P., and Datta, A.
\newblock (2018).
\newblock Gender bias in neural natural language processing.
\newblock {\em ArXiv}, abs/1807.11714.

\bibitem[\protect\citename{Mintz \bgroup et al.\egroup
  }2009]{mintz-etal-2009-distant}
Mintz, M., Bills, S., Snow, R., and Jurafsky, D.
\newblock (2009).
\newblock Distant supervision for relation extraction without labeled data.
\newblock In {\em Proceedings of the Joint Conference of the 47th Annual
  Meeting of the {ACL} and the 4th International Joint Conference on Natural
  Language Processing of the {AFNLP}}, pages 1003--1011, Suntec, Singapore,
  August. Association for Computational Linguistics.

\bibitem[\protect\citename{Papoutsaki \bgroup et al.\egroup
  }2015]{Papoutsaki2015CrowdsourcingFS}
Papoutsaki, A., Guo, H., Metaxa-Kakavouli, D., Gramazio, C., Rasley, J., Xie,
  W., Wang, G., and Huang, J.
\newblock (2015).
\newblock Crowdsourcing from scratch: A pragmatic experiment in data collection
  by novice requesters.
\newblock In {\em HCOMP}.

\bibitem[\protect\citename{Radford \bgroup et al.\egroup
  }2019]{radford2019language}
Radford, A., Wu, J., Child, R., Luan, D., Amodei, D., and Sutskever, I.
\newblock (2019).
\newblock Language models are unsupervised multitask learners.

\bibitem[\protect\citename{Rudinger \bgroup et al.\egroup
  }2018]{rudinger-etal-2018-gender}
Rudinger, R., Naradowsky, J., Leonard, B., and Van~Durme, B.
\newblock (2018).
\newblock Gender bias in coreference resolution.
\newblock In {\em Proceedings of the 2018 Conference of the North {A}merican
  Chapter of the Association for Computational Linguistics: Human Language
  Technologies, Volume 2 (Short Papers)}, pages 8--14, New Orleans, Louisiana,
  June. Association for Computational Linguistics.

\bibitem[\protect\citename{Serban \bgroup et al.\egroup }2016]{dialogue-gen}
Serban, I.~V., Sordoni, A., Bengio, Y., Courville, A., and Pineau, J.
\newblock (2016).
\newblock Building end-to-end dialogue systems using generative hierarchical
  neural network models.
\newblock In {\em Proceedings of the Thirtieth AAAI Conference on Artificial
  Intelligence}, AAAI'16, page 3776–3783. AAAI Press.

\bibitem[\protect\citename{Sun \bgroup et al.\egroup
  }2019]{sun-etal-2019-mitigating}
Sun, T., Gaut, A., Tang, S., Huang, Y., ElSherief, M., Zhao, J., Mirza, D.,
  Belding, E., Chang, K.-W., and Wang, W.~Y.
\newblock (2019).
\newblock Mitigating gender bias in natural language processing: Literature
  review.
\newblock In {\em Proceedings of the 57th Annual Meeting of the Association for
  Computational Linguistics}, pages 1630--1640, Florence, Italy, July.
  Association for Computational Linguistics.

\bibitem[\protect\citename{Yao \bgroup et al.\egroup }2019]{yao2019plan}
Yao, L., Peng, N., Ralph, W., Knight, K., Zhao, D., and Yan, R.
\newblock (2019).
\newblock Plan-and-write: Towards better automatic storytelling.
\newblock In {\em The Thirty-Third AAAI Conference on Artificial Intelligence
  (AAAI-19)}.

\bibitem[\protect\citename{Zeng \bgroup et al.\egroup
  }2015]{zeng-etal-2015-distant}
Zeng, D., Liu, K., Chen, Y., and Zhao, J.
\newblock (2015).
\newblock Distant supervision for relation extraction via piecewise
  convolutional neural networks.
\newblock In {\em Proceedings of the 2015 Conference on Empirical Methods in
  Natural Language Processing}, pages 1753--1762, Lisbon, Portugal, September.
  Association for Computational Linguistics.

\bibitem[\protect\citename{Zhao \bgroup et al.\egroup
  }2017]{zhao-etal-2017-men}
Zhao, J., Wang, T., Yatskar, M., Ordonez, V., and Chang, K.-W.
\newblock (2017).
\newblock Men also like shopping: Reducing gender bias amplification using
  corpus-level constraints.
\newblock In {\em Proceedings of the 2017 Conference on Empirical Methods in
  Natural Language Processing}, pages 2979--2989, Copenhagen, Denmark,
  September. Association for Computational Linguistics.

\bibitem[\protect\citename{Zhao \bgroup et al.\egroup
  }2018]{zhao-etal-2018-gender}
Zhao, J., Wang, T., Yatskar, M., Ordonez, V., and Chang, K.-W.
\newblock (2018).
\newblock Gender bias in coreference resolution: Evaluation and debiasing
  methods.
\newblock In {\em Proceedings of the 2018 Conference of the North {A}merican
  Chapter of the Association for Computational Linguistics: Human Language
  Technologies, Volume 2 (Short Papers)}, pages 15--20, New Orleans, Louisiana,
  June. Association for Computational Linguistics.

\bibitem[\protect\citename{Zhao \bgroup et al.\egroup
  }2019]{zhao-etal-2019-gender}
Zhao, J., Wang, T., Yatskar, M., Cotterell, R., Ordonez, V., and Chang, K.-W.
\newblock (2019).
\newblock Gender bias in contextualized word embeddings.
\newblock In {\em Proceedings of the 2019 Conference of the North {A}merican
  Chapter of the Association for Computational Linguistics: Human Language
  Technologies, Volume 1 (Long and Short Papers)}, pages 629--634, Minneapolis,
  Minnesota, June. Association for Computational Linguistics.

\end{thebibliography}

\end{document}